%% file: paper.tex
\documentclass[runningheads]{llncs}

\usepackage[T1]{fontenc}
\usepackage{graphicx}
\usepackage{hyperref}
\usepackage{color}
\usepackage{booktabs}
\usepackage{amsmath}
\usepackage{multirow}
\usepackage{amssymb}
\usepackage{tabularx}
\usepackage{array}

\begin{document}

\title{Bar-JEPA: Extracting Values from Bar Chart with Joint-Embedding Predictive Architecture}
\titlerunning{Bar-JEPA}

\author{Poonam Poonam\inst{1}
\and
Alexander Epple\inst{1}
\and Timo Ropinski\inst{1}}
\authorrunning{P. Poonam et al.}

\institute{\textsuperscript{1}\,Institute of Media Informatics, Visual Computing Group, Ulm University, Germany \\
\email{\{poonam.poonam, alexander.epple, timo.ropinski\}@uni-ulm.de}}
\maketitle

\begin{abstract}
Bar charts are commonly used in data visualization, and while they are easily understood by humans, it is non-trivial to extract the underlying data computationally. For a machine-learning-based approach, training chart de-rendering models usually requires labeled, real-world data. Labeling data is a time consuming task, which is why annotated data is scarce. Models can learn more efficiently when provided with features of high semantic quality, which a joint-embedding predictive architecture (JEPA) is designed to learn in a self-supervised manner. We present a per-bar, numerical value recovery pipeline for bar charts, where a JEPA encoder is used to produce semantically rich latent features. The decoder model consuming these features is simple and quick to train and outputs the coordinates of ticks and bars, which can be used to recover bar values. The effectiveness of self-supervised finetuning and quality of the extracted features is evident when comparing our model to end-to-end supervised baselines. Code, datasets and checkpoints are available on \href{https://github.com/dralois/Bar-JEPA}{GitHub}.

\keywords{Document analysis \and Bar chart value recovery \and Document understanding.}
\end{abstract}

\section{Introduction}
Charts are a popular tool in data visualization, as they can effectively leverage human vision to communicate the underlying data. The raw data is often lost or unavailable when charts are published, as it is easier for humans to read charts than tables~\cite{Meyer1999InformationSAA}. For machines on the other hand, having access to raw data is not only preferable, but necessary for tasks such as data analysis or for accessibility~\cite{Joshi2021UnblindTCA}. When extracting data from, or de-rendering charts, it is important for the recovered data to be both numerically and factually accurate, as well as true to the source. This is a challenging task, as there are many graphical and textual components, which need careful, task-specific handling and vary visually~\cite{huang2007system}.

Many ways have been proposed previously to recover data from charts, ranging from semi-automatic~\cite{plotdigitizerPlotDigitizerExtract}, to relying on mostly handcrafted features and methods~\cite{savva2011revision,mishchenko2011chart} and machine-learning based approaches~\cite{luo2021chartocr,cheng2023chartreader,liu2023deplot}. Deep learning methods have achieved state-of-the-art (SoTA) performance on tasks such as chart question answering, chart-to-table, and more, but require vast amounts of labeled data for training~\cite{masry2023unichart}. Synthetic data can readily be generated or accumulated~\cite{luo2021chartocr}, but the availability of real-world data, such as the CHART-Info 2024 dataset~\cite{davila2024chart}, is much more limited. The problem with synthetic data is, however, that it often does not contain artifacts and struggles to capture the full diversity of real-world data.

In this paper, we propose an approach to deal with real-world, labeled data scarcity by leveraging the self-supervised JEPA~\cite{lecun2022path}, which is trained on unlabeled data. We finetune I-JEPA~\cite{assran2023jepa}, which has been pretrained on ImageNet-1K, on bar charts to produce semantically rich feature maps. For all training tasks, we generate a synthetic bar chart dataset. Then, to test the effectiveness of JEPA for chart de-rendering, we train a set of decoders to extract per-bar numerical values. The decoders first upsample the feature maps and then regress heatmaps for the bars, ticks, and coordinate system origin. They are intentionally designed to be simple and quick to train, as it is not the main focus of the paper. Another problem we address is that aspect ratios of charts can vary significantly. For example, the bar charts in the UB PMC training dataset~\cite{davila2022icpr}, which consists of manually annotated charts, range all the way from 4.45 to 0.35. Vision transformers (ViT)~\cite{dosovitskiy2020vit}, which I-JEPA uses, require fixed-resolution inputs, so images have to either be resized or padded. To this end, we modify JEPA to accept variable-resolution inputs as proposed in Pix2Struct~\cite{lee2023pix2struct}.

To summarize the contributions of this work, we (1) propose a chart data extraction pipeline that uses I-JEPA as a feature extractor, (2) extend I-JEPA to accept variable-resolution inputs and (3) provide a synthetic, 100k bar chart dataset and accompanying generator.

\section{Related work}

Our work touches two distinct topics: Using JEPA as a feature extractor for some downstream task and chart de-rendering. Our focus, however, is only on extracting values from bar charts.

\subsection{Joint-Embedding Predictive Architectures}

Since JEPA was proposed by LeCun in 2022~\cite{lecun2022path}, it has been used for various tasks across many modalities to predict representations rather than reconstructing data. Both Fei \textit{et al.}~\cite{fei2023jepa} and more recently Tuncay \textit{et al.}~\cite{tuncay2025audio} use JEPA for audio classification. Fei \textit{et al.} employ a novel curriculum masking strategy, which they find to be crucial for downstream performance and note that the correct strategy depends on the input modality. Tuncay \textit{et al.} on the other hand, note that their training regime required far less pre-training as well as less training data. Riou \textit{et al.}~\cite{Riou2024StemJEPAAJ} use JEPA for musical stem compatibility estimation, which indicates to what degree an audio file of a single instrument matches some musical context. They do fall behind their baselines in some tasks, but note the significantly lower amount of training data they had available ($\approx 1\%$ of what the baselines used). These findings support the claim that JEPA is a more data-efficient learning framework than other self-supervised methods.

Point-JEPA by Saito \textit{et al.}~\cite{saito2025point} learns to predict point clouds, achieves SoTA performance, and the pre-training converges much faster than comparable methods. They point out that JEPA excels in environments where the available data heavily skews towards being unlabeled. Similar to charts, geometry and structure are very important in point clouds, which can be well represented by JEPA.

In the realm of vision models, I-JEPA~\cite{assran2023jepa} manages to improve upon Masked Autoencoders (MAE)~\cite{he2022masked} and Context Autoencoders (CAE)~\cite{chen2024context} in linear probing at a fraction of the compute. They highlight how I-JEPA converges faster than pixel reconstruction methods such as MAE and CAE, while learning features of a high semantic level. V-JEPA~\cite{bardes2023vjepa} and the more recent V-JEPA 2~\cite{assran2025vjepa2}, which build upon I-JEPA, can solve tasks such as spatio-temporal action detection and image classification without model parameter adaptation. Notably, V-JEPA 2 is particularly effective at tasks that require fine-grained motion understanding, such as pick-and-place on real-world robots. Overall, this shows that pixel-perfect reconstructions are not necessary to achieve good performance on many downstream tasks.

JEPA has also been used in textual contexts, for example TI-JEPA by Vo \textit{et al.}~\cite{vo2024ti} achieves SoTA performance on multimodal sentiment analysis, with the pre-training goal being multimodal alignment of images and text. They manage to bridge the semantic gap between text and image, and this is possible due to the model learning robust and generalizable features. Finally, T-JEPA by Thimonier \textit{et al.}~\cite{thimonier2024tjepa} uses JEPA as a pre-training technique for tabular data classification and regression. They manage to match or outperform gradient-boosted decision trees, which are considered to be the go-to method for tabular data.

Clearly, JEPA has been shown to be effective across a wide array of modalities, while being data efficient, learning semantic abstractions and converging fast. To the best of our knowledge it has, however, not yet been applied to charts.

\subsection{Chart De-Rendering}

Many prior works have applied neural networks to tasks such as chart de-rendering, data recovery and plot question-and-answering (QA). The approaches range from employing convolutional neural networks (CNNs) to transformer architectures and hybrid approaches.

CNNs are a natural fit for charts, as they can effectively exploit the spatial relations of images. For bar charts specifically, object detection-based methods are particularly popular. Liu \textit{et al.} use Faster-RCNN to extract text and values from bar and pie charts~\cite{liu2019dataextractnn}, while Ma \textit{et al.} rely on ResNet-50 for feature extraction and Cascade R-CNN for element detection~\cite{ma2021extractframework}. Both networks have multi-stage pipelines, chart specific heads and thus strong inductive biases. The performance suffers when evaluating real-world data compared to synthetic charts, which Liu \textit{et al.} attribute to the unavailability of labeled data. Shahira \textit{et al.} propose using Mask R-CNN for bar chart de-rendering, but also note that their approach requires a lot of training data, as well as being slow to train~\cite{shahira2023maskrcnn}. Several works have instead relied on point-based methods~\cite{hassan2023lineex,soto2023barchartppn}, where key points are extracted and further processed for value recovery. ChartOCR uses Hourglass Net to classify the chart type and extract key points from which data can be extracted, where the method depends on the chart type~\cite{luo2021chartocr}. At the time, ChartOCR achieved SoTA performance and it is easy to include additional chart types, which indicates that key point based methods are a good choice for charts. In Zhou \textit{et al.}'s work, bar charts are de-rendered by a CNN in combination with an attention mechanism and long short-term memory (LSTM)~\cite{zhou2021barchartnn}. Although the attention mechanism learns to focus on key points, the outputs are vectors containing the coordinates of bar centers and normalized heights. They also note, that their method is unlikely to generalize well to unseen data due to the lack of labeled, real-world data and the diversity thereof.

With the ever growing popularity of transformers, they have also been used in chart de-rendering, being especially popular for chart QA. 
Pix2Struct ~\cite{lee2023pix2struct}, MatCha~\cite{liu2023matcha} and DePlot~\cite{liu2023deplot} all build upon each other and use ViT~\cite{dosovitskiy2020vit} encoders and text decoders. The models achieve SoTA results on plot QA tasks and significantly outperform the CNN-based ChartOCR~\cite{luo2021chartocr}. Pix2Struct introduces variable-resolution inputs, as maintaining the original aspect ratio helps to improve model performance. There are many more examples of transformer-based chart-to-table models~\cite{cheng2023chartreader,masry2025chartgemma,masry2023unichart}, which also show promising results for chart de-rendering. Key point detection is also possible with transformers as shown by Xue \textit{et al.} in ChartDETR~\cite{xue2023chartdetr}, building upon DETR~\cite{carion2020detr}, which is an end-to-end object detector. Their method is straightforward and, while being conceptually similar, improves upon ChartOCR in terms of robustness and performance. These findings indicate that transformers can often outperform CNN-based approaches and therefore, are a good choice for the task of chart de-rendering.

It is worth pointing out, however, that even though the training objectives are diverse, all prior work, as opposed to JEPA~\cite{assran2023jepa}, fall into the field of supervised learning. To our knowledge, little to no work has been done on self-supervised pretraining in chart de-rendering. Existing methods focus on optimizing end-to-end performance and representation quality is rarely taken into account.

\section{Method}
This section covers the dataset used for training and the simplifying assumptions we make to scope the research area. As our goal is to study representation quality, we limit complexity and ensure a controlled setting. An overview of I-JEPA follows, as well as an explanation of our modifications to the framework and the self-supervised pretraining. The decoders and their supervised training regime are outlined next, before finally the numerical value extraction is detailed.

\subsection{Synthetic data generator and datasets}

\begin{figure}
\centering
\begin{minipage}{0.48\textwidth}
\centering
\includegraphics[width=\linewidth]{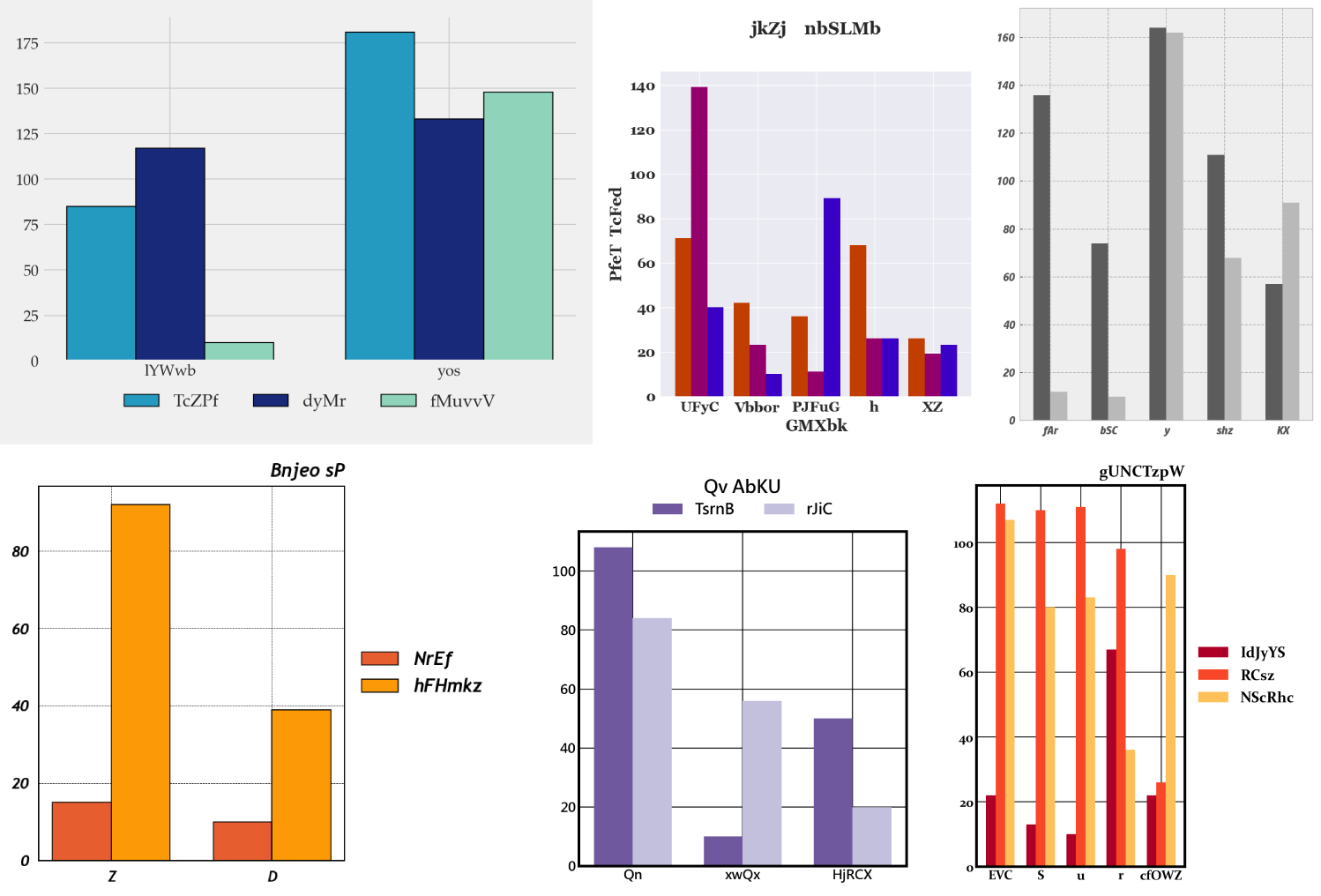}
\caption{Examples of generated bar charts as used for encoder finetuning and decoder training. Charts are generated according to the parameters in \autoref{tab:gen_params_tbl}.}
\label{fig:chart_ex}
\end{minipage}
\hfill
\begin{minipage}{0.48\textwidth}
\centering
\includegraphics[width=\linewidth]{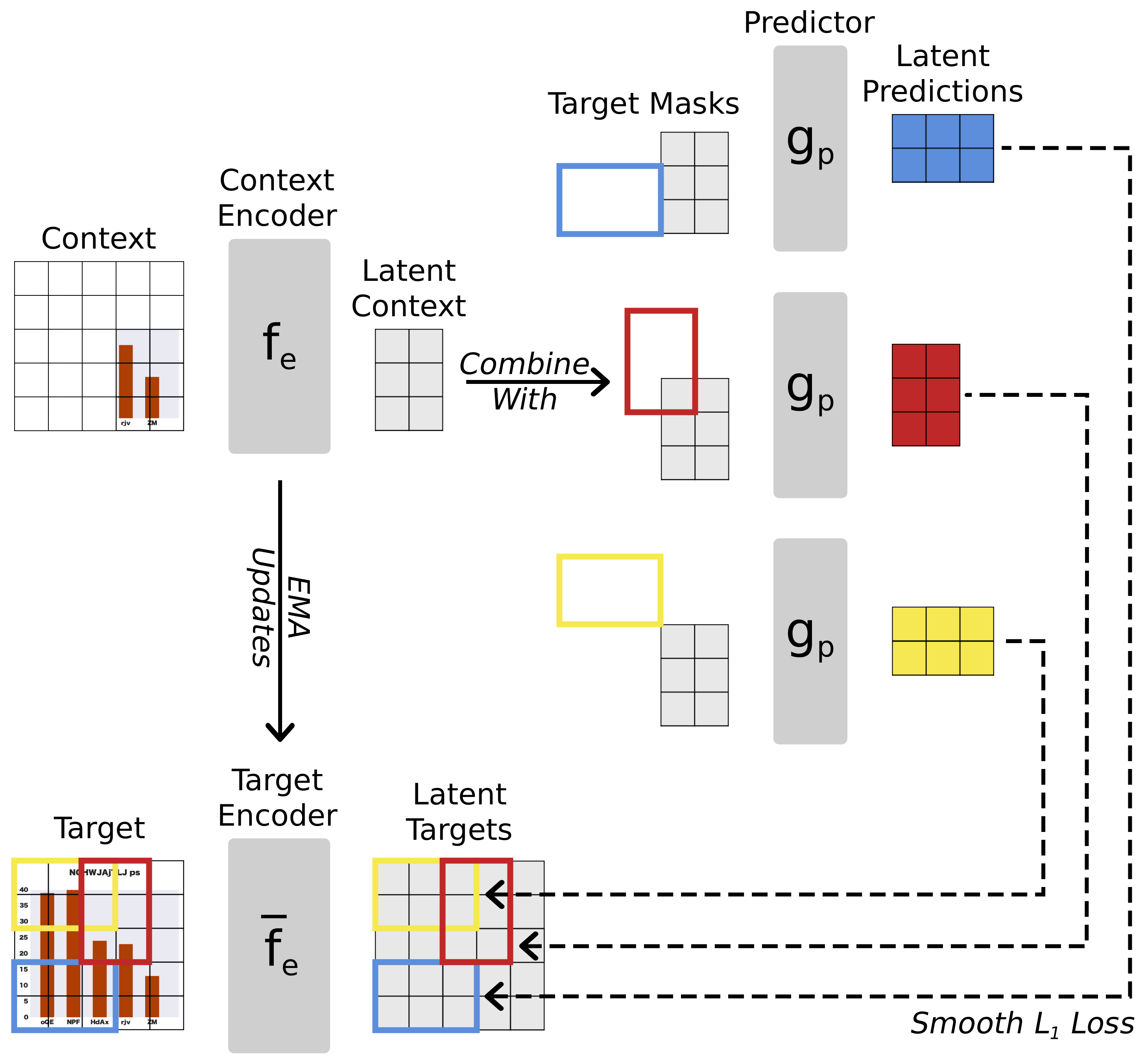}
\caption{Overview of the I-JEPA training procedure. The context encoder is used after training concludes, predictor and target encoder are discarded.}
\label{fig:jepa_ex}
\end{minipage}
\end{figure}

Bar charts come in considerable visual variety and can contain a lot of visual clutter, so in order to narrow the amount of variables, we make some assumptions for the dataset we generate for training. We only consider vertical bar charts and the bars may not be stacked, contain error bars or 3D effects. This ensures the charts have visual diversity but are kept simple. The parameters used to generate our datasets are listed in \autoref{tab:gen_params_tbl} and we use matplotlib~\cite{hunter2007mpl} to generate charts. We extend the work of Zhou \textit{et al.}~\cite{zhou2021barchartnn} by, for example, varying which random distribution is used to generate values and by outputting a more exhaustive annotation file. The annotations contain precise bounding boxes and values of all chart elements. Some examples of the charts we generate can be seen in \autoref{fig:chart_ex}. For the I-JEPA finetuning objective, we generate $100k$ charts, all of which are used for training, as there is no validation or test step. For one experiment, we include an additional finetuning step on $15k$ real-world charts from Chart-to-Text~\cite{kantharaj2022chart2text}. The dataset for the decoder pretraining consists of $17k$ synthetic charts. Additionally, we use the UB PMC dataset~\cite{davila2024chart} (specifically, ICPR CHART-Infographics 2022) for decoder finetuning. We can only use $1316$ of the vertical bar charts, as the remaining data does not have the required labels. Both datasets are split into $80:20\%$ training : validation sets.

\subsection{I-JEPA encoder}

\begin{figure}
\includegraphics[width=\textwidth]{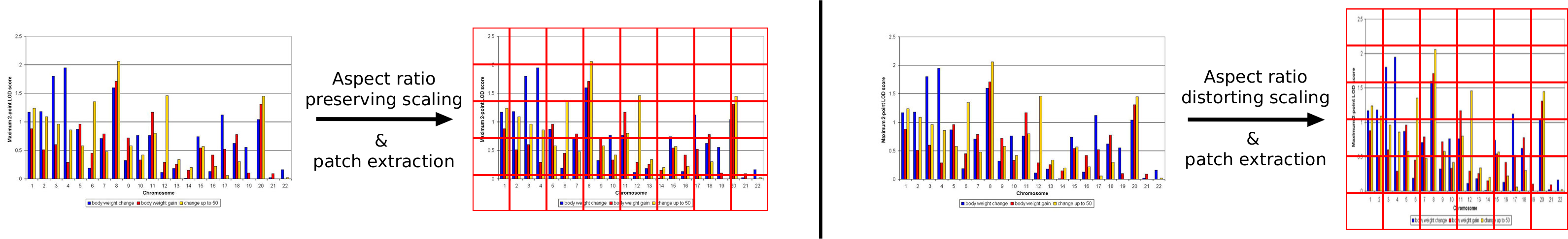}
\caption{Aspect-ratio preserving patch extraction ensures at most $N$ patches are obtained after resizing an image. The operation guarantees the new size to be close to the original aspect-ratio as much as possible. The chart is an example from~\cite{davila2022icpr}.} \label{fig:arp_pipeline}
\end{figure}

Our encoding pipeline starts with variable-resolution patch extraction, following the implementation from Pix2Struct~\cite{lee2023pix2struct}. Given a maximum number $N$ of patches, for each input image of size $(w, h)$ the amount of feasible rows and columns $(R, C)$ are computed according to \autoref{eq:feasible_rows} and \autoref{eq:feasible_columns}. Then, the image is resized to $(W, H)$ and a sequence of $M \leq N$ non-overlapping patches of size $p$ are extracted. An overview of the process can be seen in \autoref{fig:arp_pipeline}.

\begin{equation}
\begin{split}
    R = \max\!\left(1,\;\min\!\left(N,\;\left\lfloor \sqrt{N}\,\sqrt{\frac{h}{w}} \right\rfloor\right)\right),
    \qquad
    H = R\,p
    \label{eq:feasible_rows}
\end{split}
\end{equation}
\begin{equation}
\begin{split}
    C = \max\!\left(1,\;\min\!\left(N,\;\left\lfloor \sqrt{N}\,\sqrt{\frac{w}{h}} \right\rfloor\right)\right),
    \qquad
    W = C\,p
    \label{eq:feasible_columns}
\end{split}
\end{equation}

The training objective of I-JEPA is similar to that of MAEs~\cite{he2022masked}, but the key differences are that it is (1) non-generative and (2) predicts in latent as opposed to pixel space. I-JEPA consists of three ViTs~\cite{dosovitskiy2020vit}: The context encoder $f_{e}$, target encoder $\bar{f_{e}}$ and predictor $g_{p}$. The context and target encoders are structurally identical, while the predictor is narrower and shallower. Given a latent context block from an encoded image and a target mask, the predictor learns to reproduce the latent patches of the target block in the same image. The target and context blocks are obtained using multi-block masking~\cite{assran2023jepa}. For the exact parameters, see \autoref{tab:jepa_params_tbl} and refer to \autoref{fig:jepa_ex} for a visual overview of the model.

\subsubsection{Targets.}
For each image, target blocks are obtained by first sampling $i$ blocks $B_{i}$ from the $M$ input patches. The size and aspect-ratio of each $B_{i}$ is random and the blocks can overlap. They are encoded by $\bar{f_{e}}$ to yield sequences $\bar{s_{i}}$, where each sequence  $\bar{s_{i}}$ contains up to $j$ latent patches. The target encoder is frozen and receives no gradient flow during backpropagation, which prevents the encoders from collapsing. Without this, the encoders' output could simply always be a constant value, which makes predictions trivial and the representations meaningless.

\subsubsection{Context.}
There is usually only one context block $B_{x}$ of random size and aspect-ratio, where any patch regions that overlap with the target blocks $B_{i}$ are removed. The remaining patches in $B_{x}$ are encoded by $f_{e}$ into a sequence $s_{x}$ of latent patches. The context encoder receives gradients during backpropagation and updates the target encoder $\bar{f_{e}}$ via exponential moving average (EMA).

\subsubsection{Predictions.}
Instead of predicting all target blocks at once, the predictor $g_{p}$ is invoked for each target $B_{i}$ individually. It receives as input (1) the context patches $s_{x}$ and (2) learned mask tokens $t_{i}$ with added positional embeddings. The mask tokens $t_{i}$ tell the predictor where the target block $B_{i}$ is and enables it to generate the sequence $\hat{s_{i}}$ of latent patch predictions.

\subsubsection{Loss.}
After the target and prediction sequences have been obtained, the loss $\mathcal{L}$ is calculated according to \autoref{eq:jepa_loss} by accumulating the smooth L1 loss (\autoref{eq:l1_loss}) of the targets $\bar{s_{i}}$ and predictions $\hat{s_{i}}$ for each block $B_{i}$.

\begin{equation}
    \mathcal{L} = \sum_{i} \text{smoothL1}(\bar{s}_{i}, \hat{s}_{i})
    \label{eq:jepa_loss}
\end{equation}

\begin{equation}
    \text{smoothL1}(x, y) =
    \begin{cases}
    \frac{1}{2} (x - y)^2 & \text{if } |x - y| < 1 \\
    |x - y| - \frac{1}{2} & \text{otherwise}
    \end{cases}
\label{eq:l1_loss}
\end{equation}

\subsubsection{Training.}
\label{sec:jepa_training}
We use the ViT-H ImageNet-1K checkpoint from the original I-JEPA paper~\cite{assran2023jepa} and finetune for 50 epochs. The model is trained using the AdamW optimizer, with cosine weight decay between $0.02$ and $0.04$ and a cosine learning rate schedule from $5.0e-5$ to $1.0e-6$ with a 6-epoch warm-up starting at $5.0e-6$. We train both variable-resolution and fixed-resolution models on the aforementioned dataset of $100k$ generated charts using an effective batch size of 40 on two NVIDIA RTX A6000 GPUs. For one experiment, we finetune the variable-resolution checkpoint for an additional 25 epochs on $15k$ real-world charts~\cite{kantharaj2022chart2text}, with all hyperparameters halved. Similarly to Thimonier \textit{et al.}~\cite{thimonier2024tjepa}, we observe that the loss starts at a collapsed equilibrium before rising and converging. This is also the case for I-JEPA pretraining~\cite{assran2023jepa} and expected behavior.

\subsection{Key point extractor}

\begin{figure}
\includegraphics[width=\textwidth]{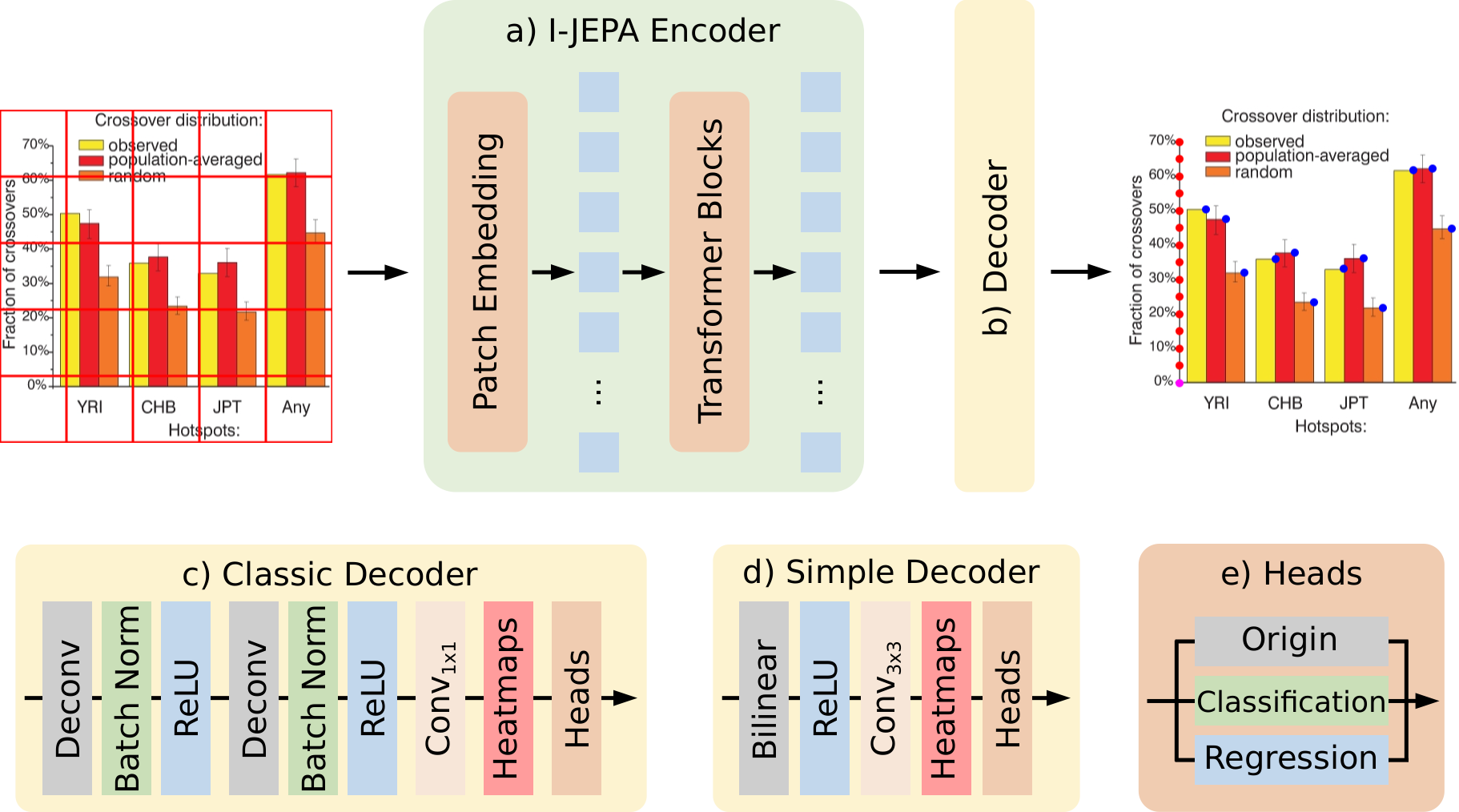}
\caption{The decoder pipeline consists of the frozen, pretrained I-JEPA encoder (a) and a decoder (b), which produces key point heatmaps. The heatmaps are consumed by the output heads (e), which in turn generate a combined heatmap. There are two possible configurations, a simple (d) and a slightly more sophisticated (c) decoder model. The chart is an example from~\cite{davila2022icpr}.}
\label{fig:decoder_pipeline}
\end{figure}

Our extraction pipeline, which can be seen in \autoref{fig:decoder_pipeline}, is intentionally simple and starts with the features extracted by the frozen encoder. These features with latent channel dimension size $d$ and patch size $p$ are of shape $F_{jepa} \in \mathbb{R}^{d \times \frac{H}{p} \times \frac{W}{p}}$. The decoder consumes these features directly and is the only part that is trained.

\subsubsection{Decoders.}
To keep the decoders lightweight, we use the simple and classic decoders described in ViTPose~\cite{NEURIPS2022_fbb10d31} but with $N_k = 32$ heatmap channels. Conceptually, this is supposed to encourage the model to put one key point into each channel during training. We enforce that the first channel contains the coordinate system origin, channels $1-15$ ticks, and the rest is reserved for bars. This is accomplished by the ground truth heatmaps being in the given order. Ticks are to be output from the bottom to the top and bars from left to right.

The simple decoder consists of a ReLU activation, followed by $4 \times$ bilinear upsampling and a $3 \times 3$ convolution. For the classic decoder, there are two blocks each consisting of a deconvolution, batch normalization and ReLU activation. The two blocks are followed by a single $1 \times 1$ convolution. The final convolutions shape the feature maps to $F_{kp}~\in~\mathbb{R}^{N_k~\times~\frac{H}{4}~\times~\frac{W}{4}}$ by reducing the channels down to $N_k$ key point maps.

\subsubsection{Heatmap heads.}
\begin{figure}
\includegraphics[width=\textwidth]{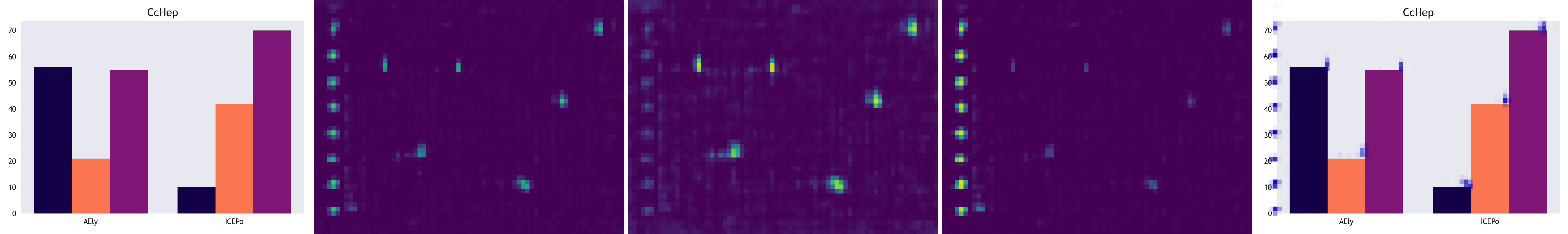}
\caption{An example of the heatmaps produced by the decoder. From left to right: Input image, (inverted) background map, top right bar corners, ticks and input image overlaid with all three heatmaps.}
\label{fig:decoder_output}
\end{figure}

The key point maps are consumed by three heads, which produce an origin, classification and regression map and mirror the method of Soto \textit{et al.}~\cite{soto2023barchartppn}. The origin map $H_{org}~\in~\mathbb{R}^{1~\times~\frac{H}{4}~\times~\frac{W}{4}}$ helps guide learning early on and predicts the location of the coordinate system origin. The channels of the classification map $H_{cls}~\in~\mathbb{R}^{3~\times~\frac{H}{4}~\times~\frac{W}{4}}$ correspond to the likelihood of a pixel being background, bar or tick. The regression map $H_{reg}~\in~\mathbb{R}^{2~\times~\frac{H}{4}~\times~\frac{W}{4}}$ accounts for point offsets accrued by the low resolution. An example of the outputs can be seen in \autoref{fig:decoder_output}.

\subsubsection{Heatmap loss.}
We use a composite loss function, consisting of separate losses for each of the three output maps and an extra loss for the key point channels. Since the decoders are trained with supervision, ground truth maps are used to compute the loss (\autoref{eq:decoder_loss}). We find through experimentation that the model is not able to output individual key point heatmaps into the key point channels directly, but the additional loss does significantly help the classification head to separate ticks from bars.

\begin{equation}
    \mathcal{L}_{total} = \lambda_1 \mathcal{L}_{org} + \lambda_2 \mathcal{L}_{cls} + \lambda_3 \mathcal{L}_{reg} + \lambda_4 \mathcal{L}_{kp}
    \label{eq:decoder_loss}
\end{equation}

\paragraph{$\mathcal{L}_{org}$:} The origin loss is computed as the MSE between a Gaussian target heatmap ($\sigma = 1.0$) and the output $H_{org}$. The origin loss is weighted equally to the classification loss ($\lambda_1 = 1.0$) and, on average, contributes $\approx25\%$ in relation to $\mathcal{L}_{cls}$ during training.

\paragraph{$\mathcal{L}_{cls}$:} For classification we use cross entropy on $H_{cls}$, with the background class being weighted $0.05$, whereas bars and ticks are weighted equally at $1.0$ due to class imbalance. Classification being most important for value recovery is weighted $\lambda_2 = 1.0$.

\paragraph{$\mathcal{L}_{reg}$:} Regression uses mean squared error (MSE) and only tick and bar pixels in $H_{reg}$ receive gradients; all other pixels are masked out. This loss is more sparse, so we set $\lambda_3 = 4.0$, but during training $\mathcal{L}_{reg}$ is, on average, only $\approx33\%$ of $\mathcal{L}_{cls}$.

\paragraph{$\mathcal{L}_{kp}$:} The heatmap point loss operates on the key point maps $F_{kp}$ and is supposed to encourage the decoder to output key points earlier, so the heads can combine them into the final outputs. As described earlier, each channel corresponds to one specific key point and its loss contribution is calculated the same as $\mathcal{L}_{org}$. Only channels with an associated ground truth key point receive gradients and the loss is normalized by the number of active channels. We weight $\lambda_4 = 4.0$, although $\mathcal{L}_{kp}$ and $\mathcal{L}_{cls}$ contribute equally to the overall loss.

\subsubsection{Training}
\label{sec:decoder_training}
The decoders are trained for 50 epochs on our generated dataset consisting of $17k$ charts. We use a cosine learning rate schedule from $1.0e-3$ to $1.0e-5$ and a 3 epoch warm-up starting at $2.0e-4$. Optimization is handled by AdamW with a cosine weight decay schedule between $0.04$ and $0.1$. The training batch size is $1360$ on a single NVIDIA RTX A6000 GPU. After pretraining concludes, we finetune for 30 more epochs on the UB PMC training dataset~\cite{davila2024chart}. Decoder finetuning follows the same regime as in \autoref{sec:jepa_training}, with a batch size of $263$ on a single NVIDIA RTX A6000 GPU, the batch size being tailored to divide the small number of samples evenly.

\subsection{Value recovery}

To translate the heatmaps into useful data, we follow the method of Soto \textit{et al.}~\cite{soto2023barchartppn} with some modifications: First, the bar and tick maps from $H_{cls}$ are background masked at a threshold of $0.75$ and surviving candidates from each map with confidence $c \geq 0.75$ are transformed to normalized image coordinates. The offsets from $H_{reg}$ are added to allow for sub-pixel accuracy. Next, confidence-based non-maximum suppression (NMS) is applied to the candidates, where the radius $r_{\text{nms}}$ can account for variable-resolution inputs and is the equivalent distance of $1.5$ pixels in heatmap size (\autoref{eq:nms_radius}). Afterwards, the remaining coordinates are the predicted bar and tick positions in normalized image space.

\begin{equation}
    r_{\text{nms}} = \frac{1.5 \cdot p}{4 \cdot\sqrt{HW}}
    \label{eq:nms_radius}
\end{equation}

The next step consists of converting tick labels to numerical data. As optical character recognition (OCR) is not the focus of this work, we use PaddleOCR ~\cite{cui2025paddleocr} for this task, specifically the \href{https://www.paddleocr.ai/latest/en/version3.x/algorithm/PP-OCRv5/PP-OCRv5_multi_languages.html#3-performance-comparison}{\textit{latin\_PP-OCRv5\_mobile\_rec}} text recognition model with the language set to English. For all detections containing numerical text (determined using regular expressions), the extracted values and bounding box centers are matched with the predicted tick positions using the Hungarian algorithm~\cite{kuhn1955hungarian}. A match is only considered to be valid if the distance $d$ between label and tick is $d \leq 5.0 \times r_{\text{nms}}$ to avoid numerical text that does not stem from labels being matched with ticks. Similar to Zhou \textit{et al.}~\cite{zhou2021barchartnn}, we fit a line for matched value and tick coordinates with RANSAC regression~\cite{fischler1981random}. The resulting function can then be used to predict values from bar coordinates.

\section{Experiments}
This section outlines the metrics used for evaluation, shows the effectiveness of finetuning the encoder on domain-specific data, and explains the influence of variable-resolution inputs, as well as the decoder choice has on performance. Finally, a short comparison to prior work highlights the strengths and shortcomings of our approach.

\begin{table}
\caption{Quantitative results of six models evaluated on two datasets. The variable- and fixed-resolution models, as well as the vanilla model, use the \textit{classic} decoder from \autoref{fig:decoder_pipeline} and the \textit{simple} decoder receives variable-resolution inputs. \textit{Real-world FT} and \textit{Oracle OCR} are finetuned versions of the variable-resolution model.}
\input{tables/eval_results}
\label{tab:eval_results}
\end{table}

\subsection{Metrics and test datasets}
All results are evaluated on two datasets, namely $100$ synthetic charts that we generate specifically for testing and split 4 of the UB PMC~\cite{davila2024chart} test set (ICPR CHART-Infographics 2022). There are only $139$ vertical bar charts in this split that have the labels required for evaluation. The results we report are the F1 score for bar and tick detections, as well as the value recovery accuracy.

To calculate the F1 score, we use the Hungarian algorithm~\cite{kuhn1955hungarian} with the precision threshold $r_{\text{f1}} = r_{\text{nms}}$ (\autoref{eq:nms_radius}) to match predicted and ground truth coordinates. When evaluating fixed-resolution inputs, the heatmaps are $64 \times 64$ pixels, so the error of a correct prediction is at most $\varepsilon \approx 2.3\%$. For consistency, we use the same threshold for both variable and fixed-resolution inputs, even though this could, in some cases, mean a fixed-resolution detection would not be considered valid in a variable-resolution setup and vice-versa.

Value recovery is measured using accuracy. We specifically use the same criterion as Zhou \textit{et al.}~\cite{zhou2021barchartnn} as described in \autoref{eq:acc_metric}. 

\begin{equation}
    \frac{\lvert h_g - h_p \rvert}{h_g} \leq \varepsilon
    \label{eq:acc_metric}
\end{equation}

Here, $h_g$ is the ground truth value, $h_p$ the predicted value and $\varepsilon$ the strictness. For our results, we use the same relaxed ($\varepsilon = 0.05$) and hard ($\varepsilon = 0.02$) accuracy thresholds. We also account for missing and additional predictions, but for simplicity no OCR correction is applied. In an oracle OCR experiment, where detected labels were replaced with ground truths, we verified that OCR is not detrimental to prediction accuracy (see \autoref{tab:eval_results}).

\subsection{Ablation study and analysis}

In order to evaluate the influence of our contributions and implementation choices, we train four different models as described in \autoref{tab:models} according to the schedule outlined in \autoref{sec:decoder_training}.

\begin{table}
\centering
\caption{The models used for the ablation studies. All encoders originate from the ViT-H ImageNet-1K checkpoint. For aspect ratio preserving (ARP) models the encoder was finetuned with ARP enabled. The decoder types are described in \autoref{fig:decoder_pipeline}.}
\input{tables/models}
\label{tab:models}
\end{table}

\subsubsection{The influence of encoder finetuning.}
Arguably, the most important question was whether or not self-supervised encoder finetuning on charts has an impact on downstream performance. As is evident from the results in \autoref{tab:eval_results}, this very much appears to be the case: On synthetic charts, the F1 scores of the fixed-resolution, finetuned encoder are more than double that of the vanilla encoder. In case of value recovery, the difference is even more pronounced. This likely is the result of compounding errors, as it depends on both bars and ticks being accurately detected. When evaluating the models on real-world data, the vanilla encoder completely fails at recovering any values, whereas the finetuned encoder has a relaxed accuracy of $42\%$. With the only difference between these two models being the additional training step, self-supervised finetuning on charts clearly is able to yield a significantly more capable encoder for downstream tasks. Additionally, the method continues to scale with more data, as finetuning the encoder further on real-world data yields even better results (e.g., $5\%$ better relaxed accuracy on the real-world test set).

\subsubsection{The influence of variable-resolution inputs.}
The second question we want to answer concerns the impact of variable-resolution inputs. In Pix2Struct~\cite{lee2023pix2struct}, this simple change resulted in $\approx 5\%$ accuracy uplift in their warm-up stage. We see a similar benefit in value recovery accuracy, the difference being $\approx 5\%$ for real world data and $\approx 13\%$ for synthetic charts at the hard accuracy threshold. Considering this modification comes at negligible cost during training and inference time, the integration is well worth it.

\subsubsection{The influence of decoder choice.}
Finally, we find that the results of ViTPose~\cite{NEURIPS2022_fbb10d31} do not transfer to our method. Whereas they see almost no performance difference between the two decoders for human pose estimation, the simple decoder performs significantly worse for our task. It fails to recover values from both synthetic and real-world charts, with the accuracy hovering at $\approx 1\%$. When evaluating the heatmaps directly, the outputs are diffuse and poorly localized compared to the classic decoder, which produces sharp, focused activations (Figure~\ref{fig:simple_vs_classic}). Our hypothesis is that the decoder is over-constrained and does not have enough capacity for the task at hand, which is supported by the fact that instead of confusing bars and ticks, detections are simply missed (Figure~\ref{fig:confusion_matrix}). Our network is not trained end-to-end, which could possibly explain the different findings: If the encoder in ViTPose does a lot of the heavy-lifting, the decoder choice would matter a lot less.

\subsection{Comparison to other papers}

\begin{table}
\centering
\begin{minipage}[t]{0.48\textwidth}
\caption{The mean F1 score of our model and Soto \textit{et al}'s~\cite{soto2023barchartppn} trained on synthetic data. The test datasets, while not being the same, are very similar, and our precision threshold is also slightly stricter ($2.3\%$ vs. $2.7\%$).}
\input{tables/vs_soto}
\label{tab:vs_soto}
\end{minipage}
\hfill
\begin{minipage}[t]{0.48\textwidth}
\caption{Value recovery accuracy at the relaxed threshold ($\varepsilon=0.05$) of our model and Zhou \textit{et al.}'s~\cite{zhou2021barchartnn} without OCR correction. The test datasets differ. Our real-world data is more varied, whereas the synthetic data is nearly identical.}
\input{tables/vs_zhou}
\label{tab:vs_zhou}
\end{minipage}
\end{table}

It is somewhat difficult to compare our results to prior work directly, as our outputs are limited to bar and tick coordinates. Nevertheless, we can draw a comparison to the works of Soto \textit{et al.}~\cite{soto2023barchartppn} and Zhou \textit{et al.}~\cite{zhou2021barchartnn}. We can only compare results conceptually on qualitatively similar data due to the unavailability of the datasets they used in testing. As can be seen in \autoref{tab:vs_zhou}, our method performs similarly well on synthetic data at the relaxed threshold ($\varepsilon=0.05$). Since our chart generator builds on theirs, the comparison on synthetic data is fair. While our method does fall behind on real-world data, most of the charts in our test set do not fulfill the criteria Zhou \textit{et al.} assume (e.g. include error bars / stacked bars). To compare our work to Soto \textit{et al.}, we use the mean tick and bar F1 scores over the entire test set. Their synthetic data is, again, very comparable to ours and the real-world data they curated, just like UB PMC~\cite{davila2024chart}, consists of charts from PubMedCentral. The results in \autoref{tab:vs_soto} show that, while our model performs slightly worse on synthetic data ($\approx 0.03$ difference), we can improve upon their results when it comes to real-world data by $\approx 0.26$. It should be noted, however, that we do not have many trainable parameters in the value recovery pipeline, as the bulk of the parameters live in the frozen encoder. Additionally, Zhou \textit{et al.}~\cite{zhou2021barchartnn} train on $30k$ charts for $300$ epochs and Soto \textit{et al.} $1750$ epochs on $5k$ charts, which is significantly more than it takes for our model to converge (see \autoref{sec:decoder_training}). This suggests that self-supervised finetuning, as well as the semantic quality of the extracted features are the driving factor behind our results.

\section{Discussion and Conclusion}
In this study, we propose utilizing I-JEPA, a self-supervised learning model, as a feature extractor for chart de-rendering. To our knowledge, this is the first time that JEPA has been used in the context of chart understanding, a traditionally supervision-heavy task. We are able to show that finetuning the feature extractor is highly beneficial for downstream performance and our results suggest that the features are indeed of high semantic quality. This is supported by the fact that our lightweight value recovery model converges quickly during training and does not require a large corpus to achieve good results. We also introduce variable resolution inputs to I-JEPA, which yields modest performance improvements with negligible overhead.

Our work does, however, have some limitations: (1) Our model is rather simplistic and can only recover values of one type of chart, (2) it does not achieve SoTA results at this task and (3) it may prove difficult to integrate our feature extractor into multimodal language models (e.g. for chart QA). Although we are not trying to compete with SoTA models, it would be beneficial to build a more competent decoder in future work to show the full potential of our method. Our encoder is frozen during training and I-JEPA operates exclusively in latent space, so aligning a text-based model could be challenging. Considering many SoTA models are to some extent based on large language models~\cite{liu2023deplot,cheng2023chartreader,masry2025chartgemma,masry2023unichart}, the ability to train end-to-end might be beneficial.

There are several possible avenues for future work. Implementing a more powerful decoder, such as the transformer-based ChartDETR~\cite{xue2023chartdetr} could result in better de-rendering capabilities. Including a large language model in the pipeline or replacing the decoder with a large language model would allow testing if our method is also suited for chart QA. Taking ChartGemma~\cite{masry2025chartgemma} as an example, it may be possible to use I-JEPA as the vision encoder and align it with the language model by training an embedding layer. Finally, scaling the model to more chart types and finetuning on a more diverse corpus would be a logical next step, as it is unclear how much data is required for finetuning. Judging the quality of the extracted features directly is a hard task by itself and it is entirely possible that more data and longer training would yield even stronger features.

\begin{credits}
\subsubsection{\ackname}
We acknowledge the EuroHPC Joint Undertaking for awarding this project access to the EuroHPC supercomputer LEONARDO, hosted by CINECA (Italy) and the LEONARDO consortium through an EuroHPC Development Access call. The authors acknowledge support by the state of Baden-Württemberg through bwHPC. We also thank Laurna Epple and Hannah Kniesel for proofreading and valuable discussions.
\end{credits}
\bibliographystyle{splncs04}
\bibliography{bibliography}

\begin{appendix}
\section{Parameters}
\begin{table}
\centering
\begin{minipage}[t]{0.48\textwidth}
\caption{Parameters used to generate our datasets. The colors are selected to be rich in contrast, to avoid white-on-white charts. Fonts have to support the roman alphabet and we filter out specialty fonts (e.g. math, pictographs).}
\input{tables/gen_params.tex}
\label{tab:gen_params_tbl}
\end{minipage}
\hfill
\begin{minipage}[t]{0.48\textwidth}
\centering
\caption{I-JEPA finetuning parameters. We use the ViT-H preset for training and the default multiblock sampling strategy, with the same settings as used during pretraining. We generally allow at most 256 patches of size 14 px.}
\input{tables/jepa_params.tex}
\label{tab:jepa_params_tbl}
\vfill
\vspace{0.1cm}
\includegraphics[width=0.75\linewidth]{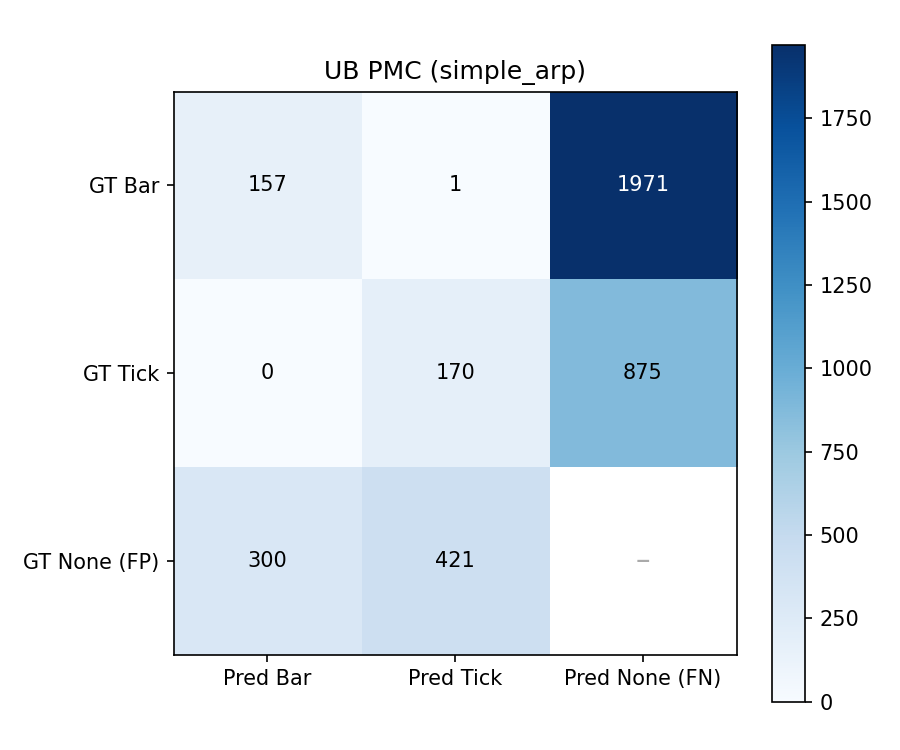}
\smallskip
\refstepcounter{figure}
\begin{center}
\parbox{0.85\linewidth}{
\centering
\textbf{Fig.~\thefigure.} The confusion matrix of the simple decoder evaluated on \cite{davila2022icpr}, showing that most bars and ticks are not detected.
}
\end{center}
\label{fig:confusion_matrix}
\end{minipage}
\begin{minipage}[t]{\textwidth}
\includegraphics[width=\linewidth]{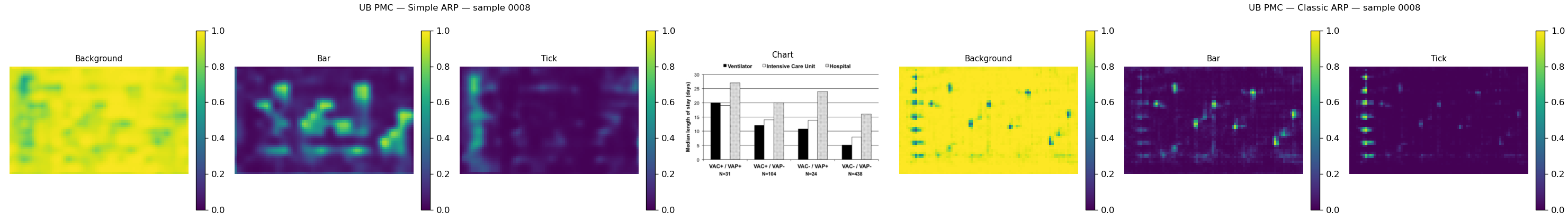}
\smallskip
\refstepcounter{figure}
\noindent
\textbf{Fig.~\thefigure.} Simple vs. classic decoder on a chart from \cite{davila2022icpr}. The activations of the simple decoder are blurry, whereas the classic decoder produces pronounced, sharp peaks.
\label{fig:simple_vs_classic}
\end{minipage}
\end{table}

\end{appendix}

\end{document}

%% file: tables/eval_results.tex
\begin{tabular}{|l|l|l|c|c|c|c|}
\hline
Dataset                  & Model                   & Bar F1 & Tick F1 & Acc. ($\varepsilon = 0.05$) & Acc. ($\varepsilon = 0.02$) \\
\hline
\multirow{4}{*}{UB PMC~\cite{davila2024chart}}
                         & Oracle OCR              & 0.785  & 0.842   & 0.512        & 0.389      \\
                         & Real-world FT           & 0.785  & 0.842   & 0.499        & 0.365      \\
                         & Variable Resolution     & 0.740  & 0.827   & 0.450        & 0.341      \\
                         & Fixed Resolution        & 0.716  & 0.812   & 0.422        & 0.292      \\
                         & Vanilla                 & 0.161  & 0.024   & 0.000        & 0.000      \\
                         & Simple Decoder          & 0.085  & 0.178   & 0.009        & 0.005      \\
\hline
\multirow{4}{*}{Synthetic Charts}
                         & Oracle OCR              & 0.961  & 0.951   & 0.783        & 0.652      \\
                         & Real-world FT           & 0.961  & 0.951   & 0.792        & 0.657      \\
                         & Variable Resolution     & 0.956  & 0.940   & 0.778        & 0.629      \\
                         & Fixed Resolution        & 0.882  & 0.892   & 0.654        & 0.498      \\
                         & Vanilla                 & 0.389  & 0.182   & 0.056        & 0.051      \\
                         & Simple Decoder          & 0.124  & 0.253   & 0.014        & 0.006      \\
\hline
\end{tabular}

%% file: tables/models.tex
\begin{tabular}{|l|c|c|c|}
\hline
Model               & ARP & Decoder & Encoder               \\
\hline
Real-world FT \& Oracle OCR       & \checkmark              & Classic & Real-world Finetuned  \\
Variable Resolution & \checkmark              & Classic & Finetuned             \\
Fixed Resolution    & $\times$                & Classic & Finetuned             \\
Vanilla             & $\times$                & Classic & ViT-H ImageNet-1K     \\
Simple Decoder      & \checkmark              & Simple  & Finetuned             \\
\hline
\end{tabular}

%% file: tables/vs_soto.tex
\begin{tabular}{|l|l|c|}
\hline
Method                                          & Data         & F1 Score \\
\hline
Ours                                            & Synthetic    & 0.956    \\
Soto \textit{et al.}~\cite{soto2023barchartppn} & Synthetic    & \textbf{0.981}    \\
\hline
Ours                                            & Real World   & \textbf{0.814}    \\
Soto \textit{et al.}~\cite{soto2023barchartppn} & Real World   & 0.555    \\
\hline
\end{tabular}

%% file: tables/vs_zhou.tex
\begin{tabular}{|l|l|c|}
\hline
Method                                         & Dataset      & Accuracy \\
\hline
Ours                                           & Synthetic    & 79\%     \\
Zhou \textit{et al.}~\cite{zhou2021barchartnn} & Synthetic    & \textbf{82\%}     \\
\hline
Ours                                           & Real World   & 50\%     \\
Zhou \textit{et al.}~\cite{zhou2021barchartnn} & Real World   & \textbf{71\%}     \\
\hline
\end{tabular}

%% file: tables/gen_params.tex

\renewcommand{\arraystretch}{1.1}
\begin{tabularx}{\linewidth}{|>{\raggedright\arraybackslash}p{0.42\linewidth}|>{\raggedright\arraybackslash}X|}
\hline
\textbf{Parameter} & \textbf{Details} \\
\hline
Chart size          & 600--1600 px (width \& height) \\
Bar series          & 2--5 \\
Bars per series     & 1--3 \\
Bar values          & 10--200 \\
Bar width \& spacing & Random in accordance with size \\
Bar colors          & Randomly sampled from color maps \\
Font \& font size   & Randomly selected from roman fonts \\
Axis label length   & 5--15 characters \\
Ticks label length  & 1--5 characters \\
Title length        & 5--15 characters \\
Title location      & left, center, right \\
Legend length       & 3--6 characters \\
Legend location     & top, bottom, right \\
\hline
\end{tabularx}
\renewcommand{\arraystretch}{1}

%% file: tables/jepa_params.tex
\begin{tabular}{|l|l|}
\hline
\multicolumn{2}{|c|}{\textbf{Context Encoder}} \\
\hline
Max Patch Count    & 256       \\
Block Count        & 1         \\
Block Size         & 85--100\% \\
\hline
\multicolumn{2}{|c|}{\textbf{Predictor}} \\
\hline
Block Aspect Ratio & 0.75--1.5 \\
Block Count        & 4         \\
Block Size         & 15--20\%  \\
\hline
\end{tabular}